\pdfoutput=1
\documentclass[10pt,twocolumn,letterpaper]{article}
\usepackage{coawm-preprint}
\usepackage{amsmath,amssymb}

\definecolor{linkblue}{rgb}{0.21,0.49,0.74}
\usepackage[pagebackref,breaklinks,colorlinks,allcolors=linkblue]{hyperref}
\title{Continuous-Action World Models: Learning Cross-Embodiment Spatiotemporal Dynamics Priors}
\author{Liangyu Li\thanks{\raggedright Tongji University. Emails: \mbox{\texttt{2550703@tongji.edu.cn}} and \mbox{\texttt{lcc4321@tongji.edu.cn}}.}\qquad Qingwen Liu\thanks{\raggedright Corresponding author. Tongji University. Email: \mbox{\texttt{qliu@tongji.edu.cn}}.}}
\date{}
\hypersetup{pdftitle={Continuous-Action World Models: Learning Cross-Embodiment Spatiotemporal Dynamics Priors},pdfauthor={Liangyu Li and Qingwen Liu}}
\begin{document}
\maketitle
\begin{abstract}
We introduce Continuous-Action World Models (CoAWM), a framework for learning cross-embodiment spatiotemporal dynamics priors that support robot action generation across embodiments and execution frequencies. At its core is a shared latent function space that captures the spatial and temporal structure of robot actions. We learn this space through two stages. First, function pretraining brings trajectories from different embodiments and sampling frequencies into a common representation, learning shared dynamics priors through action reconstruction and consistency across temporal sampling views. By accommodating heterogeneous robot trajectories, this stage broadens the range of data sources available for pretraining and facilitates scaling to larger and more diverse datasets. Second, we connect the pretrained function space to a world-model backbone, enabling observations, language instructions, and the current robot state to condition the prediction of a latent action function. A single decoder takes any query time and returns the corresponding end-effector action. By separating shared dynamics priors from task-conditioned action selection and temporal sampling, our approach provides a unified foundation for transferring robot experience across embodiments and control frequencies.
\end{abstract}
\section{Introduction}
\begin{figure*}[!tp]
\centering
\includegraphics[width=\textwidth]{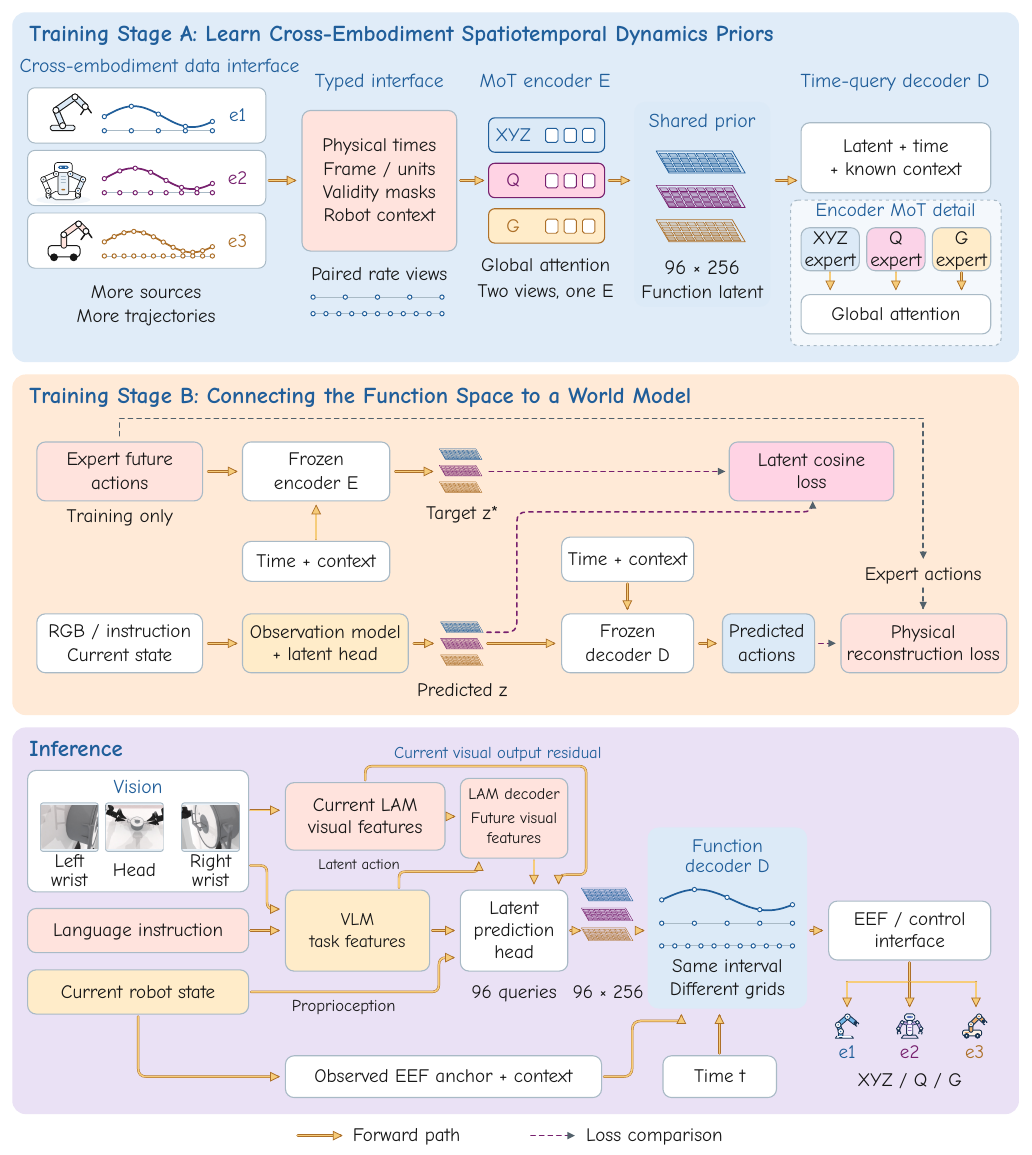}
\caption{Overview of continuous action-function learning and deployment. Training Stage A learns shared spatiotemporal priors from typed trajectories across embodiments and temporal sampling grids. Training Stage B connects observation-conditioned latent prediction to the learned function space using latent and physical reconstruction supervision. At inference, current multimodal observations and visual features condition latent prediction; the function decoder combines the predicted latent with physical query times and observed context to produce embodiment-specific control commands.}
\label{fig:interface}
\end{figure*}
Robots with different bodies can perform similar tasks, yet their experience is recorded through different action interfaces and control frequencies~\cite{oxe2023,ghosh2024octo,black2024pi0}. A reaching motion may be demonstrated by one manipulator and executed by another, with different command sequences governing a related physical behavior. This diversity raises a central question for robot world models: how can experience collected across embodiments be used to learn shared spatiotemporal dynamics priors that guide action generation? Answering this question would broaden the sources of experience available to a model, allowing robot learning to benefit from data beyond a single embodiment or collection setup.

The challenge is not simply to combine more trajectories. Robot datasets differ in what their actions specify, how those actions are grounded in the robot's state, and when commands are issued. The same sequence index can correspond to different physical times, while numerically similar commands can have different meanings across action interfaces. A common storage format alone does not resolve these differences. To learn from heterogeneous experience, a model needs a representation that captures reusable spatial and temporal structure while retaining the information required to interpret and execute an action on a particular robot. For a world model, this representation must also connect shared experience to action selection in the current scene and task.

\begin{figure*}[t]
\centering
\includegraphics[width=\textwidth]{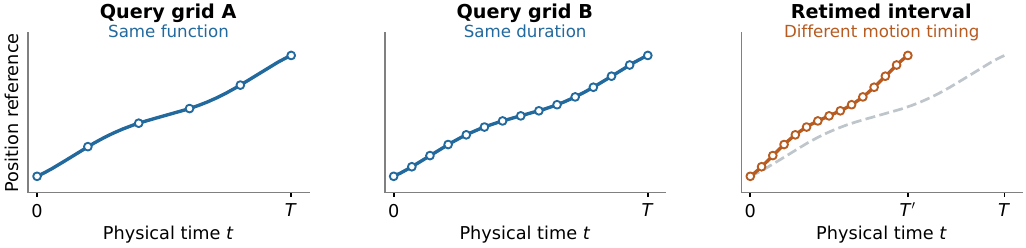}
\caption{\textbf{An action interval and its sampling grid are different objects.} Schematic of a continuous scalar position reference. The first two panels query the same function over the same physical duration with different grids. The third changes the duration: the geometric progression is preserved, but its velocity and acceleration change. A reusable function representation must retain physical timing while allowing the execution grid to vary.}
\label{fig:insight}
\end{figure*}

Our key insight is to learn the evolution of an action as a function, rather than treating its sampled commands as the fundamental object of representation. A latent action function encodes how a motion unfolds in space and time; individual commands are obtained by querying that function (Fig.~\ref{fig:insight}). This provides a common learning space for trajectories sampled at different frequencies. More importantly, it separates two complementary learning problems: acquiring dynamics priors from heterogeneous trajectories, and selecting an action using the current interaction context. The former can draw on a broader pool of robot experience, while the latter connects these priors to the observation-conditioned capabilities of a world model.

We introduce Continuous-Action World Models (CoAWM) to realize this idea through two stages (Fig.~\ref{fig:interface}). First, we pretrain a shared latent function space using trajectories from different embodiments and sampling frequencies. A function encoder and decoder learn spatiotemporal dynamics priors through action reconstruction and consistency across temporal sampling views. This pretraining stage expands the range of trajectory sources that can contribute to representation learning, providing a path toward larger and more diverse training datasets. Second, we connect the pretrained function space to a world-model backbone. Visual observations, language instructions, and the current robot state condition a prediction head that produces a latent action function, which a shared decoder converts into end-effector actions through time queries. CoAWM thus connects cross-embodiment prior learning with task-conditioned action generation, while allowing the execution frequency to be specified through the query times.

\section{Related Work}

\subsection{Continuous Action and Function Representations}

Action representations determine how a policy organizes motion over time. ACT predicts action chunks~\cite{zhao2023act}, Diffusion Policy generates action sequences through denoising~\cite{chi2023diffusionpolicy}, and FAST compresses temporal structure into frequency-domain tokens~\cite{pertsch2025fast}. Continuous action representations capture how motion evolves over time and allow actions to be queried at different execution times. Dynamical Movement Primitives encode motor behavior through attractor dynamics and learned forcing terms~\cite{ijspeert2013dmp}, Probabilistic Movement Primitives model distributions over trajectories~\cite{paraschos2013promp}, and ProDMPs connect these dynamic and probabilistic formulations~\cite{li2022prodmp}. These foundations establish motion trajectories as structured objects from which reusable action priors can be learned.

Recent visuomotor policies explore different function families and decoding mechanisms. FLASH represents trajectories with a Legendre basis and history-anchored flow~\cite{bai2026flash}, while B-spline Policy uses control points and knots to parameterize actions~\cite{han2026bspline}. Neural Implicit Action Fields learn modulated continuous-time action functions and supervise their temporal derivatives~\cite{liu2026niaf}. Periodic-activation implicit networks provide a general foundation for coordinate-queryable neural functions and their derivatives~\cite{sitzmann2020siren}. These efforts to represent actions motivate us to learn shared spatiotemporal dynamics priors through a continuous function representation.

A latent function representation captures how motion evolves over time. A decoder takes this representation and a time query to produce the corresponding action. Querying the same representation at different times yields action sequences at different temporal resolutions. FunDiff combines function encoding, latent diffusion, and continuous decoding for physical-field generation~\cite{wang2026fundiff}. Function-Space Diffusion for Motion Planning models trajectory distributions using temporally correlated noise and boundary-compatible operators~\cite{chang2026fsdmp}. CAT learns a fixed-size set of continuous trajectory latents over a physical-time window and uses frequency-aware positional encoding with a causal action decoder to accommodate different control rates~\cite{yang2026cat}. These works establish important precedents for learning continuous representations and transferring between temporal resolutions. CoAWM focuses on learning shared spatiotemporal dynamics priors from heterogeneous robot trajectories and connecting the resulting function space to world-model-conditioned action generation.

\subsection{Cross-Embodiment Learning and Action Priors}

Cross-embodiment learning expands the experience available for robot policy training. Open X-Embodiment brings together robot datasets and studies cross-robot policies~\cite{oxe2023}. Octo develops generalist pretraining and adaptation to new observation and action spaces~\cite{ghosh2024octo}, and $\pi_0$ combines a vision-language model with flow-based action generation for general robot control~\cite{black2024pi0}. These approaches make heterogeneous data a practical resource for learning behavior across tasks and robot systems. A central representation question is which structure can be shared while maintaining a meaningful connection to each robot's executable actions.

Latent-action learning addresses this question by separating reusable action knowledge from specific command spaces. LAPA learns latent actions from video before adapting them to robot control~\cite{ye2024lapa}, and UniVLA learns task-centric latent actions for embodiment transfer~\cite{bu2025univla}. LAC-WM learns shared latent actions from visual transitions with auxiliary motion reconstruction for cross-embodiment world modeling, and reports benefits from increasing the diversity of pretraining embodiments~\cite{huang2026lacwm}. These results motivate shared action priors for both control and prediction. CoAWM approaches this problem through action trajectories: a function autoencoder learns their spatial and temporal structure, and an observation-conditioned head connects the pretrained function space to a world-model backbone. Its pretraining interface is designed to accommodate trajectories from multiple embodiments and sampling frequencies, broadening the sources of data available for learning dynamics priors.

\subsection{World Models for Action Generation and Execution}

World models connect experience to predictions that support action selection. V-JEPA 2 adapts self-supervised video representations for action-conditioned prediction and planning~\cite{assran2025vjepa2}. LAWM learns latent actions through future-frame modeling~\cite{tharwat2025lawm}, while Learning Latent Action World Models In The Wild investigates continuous-valued latent actions for prediction and controllable interaction~\cite{garrido2026wild}. villa-X combines latent-action modeling with visual and proprioceptive forward dynamics and robot-action prediction~\cite{chen2025villax}. Alongside LAC-WM~\cite{huang2026lacwm}, these approaches highlight the role of an action representation in linking predicted transitions to executable behavior.

LaWAM conditions action generation on predicted latent visual subgoals and explicitly supports physical-time alignment and mixed-frequency training~\cite{chen2026lawam}. Its mixed-frequency setting retains native-rate action sequences with variable lengths and uses physical-time encoding to align them. AHA-WAM separates a slow video planner from a fast action executor through observation-guided context routing and horizon-adaptive offset training~\cite{cai2026ahawam}. AW-0 extends world-action modeling with joint code, video, and action generation~\cite{liu2026aw0}. CoAWM learns a shared latent function space through cross-embodiment pretraining and connects it to world model and action generation. Given visual observations, language instructions, and the current robot state, the model predicts a latent function in this pretrained space. The decoder then takes time queries to produce executable actions. This design makes the pretrained dynamics priors directly available to action generation and allows a single predicted function to produce action sequences at different execution frequencies.

Execution introduces complementary questions about continuity, latency, and commitment. Real-Time Chunking overlaps prediction and execution through inference-time inpainting~\cite{black2025rtc}, Legato learns native continuation for action-chunking flow policies~\cite{liu2026legato}, and REMAC uses masked action chunking to address real-time execution~\cite{wang2026remac}. Mixture of Horizons combines action predictions over multiple horizons~\cite{jing2025moh}; PACE uses kinematic phases to select execution boundaries~\cite{nie2026pace}, and DVAC adapts the executed prefix using denoising variance~\cite{feng2026dvac}. Reliable Action-Chunk Extension models when to switch according to predicted subskill transitions~\cite{yu2026raceextension}, whereas Reachability-aware Accelerated Chunk Execution retimes predicted state paths under physical constraints~\cite{kim2026racetimeoptimal}. These methods concern how a plan is committed, continued, or retimed during execution. CoAWM complements these execution mechanisms by separating the latent action function, the times at which it is queried, and the physical timing of the represented motion.

\section{Method}
\label{sec:method}
\subsection{Overview}\label{overview}

Continuous-Action World Models (CoAWM) connect world model and action generation through a pretrained space of action functions. The central representation is a latent function that describes how an action evolves over physical time. A decoder takes this latent function and time queries to produce the corresponding actions, enabling action generation at different execution frequencies through queries at the corresponding time points.

CoAWM uses two training stages. Stage A learns an action-function encoder and decoder from timestamped robot trajectories. A Mixture-of-Transformers (MoT) encoder combines specialized processing of end-effector position, orientation, and gripper commands with joint attention across these action components. Training aligns representations of the same motion observed at different frequencies. This provides a shared function space for learning from trajectories collected across robot embodiments and at different frequencies, broadening the data available for pretraining. Stage B connects the learned function space to a world-model backbone. The prediction head combines VLM features, predicted future visual features, and robot proprioception, with current visual features contributing through an output residual. It predicts a function in the pretrained space. Expert trajectories supervise both its latent representation and the actions obtained by decoding it. At inference, the decoder queries the predicted latent function at time points determined by the desired execution frequency.

\subsection{Continuous Action Function Representation}\label{continuous-action-function-representation}\label{sec:representation}

\textbf{A function latent with a physical-time interface.} We represent an action sequence as timestamped observations \(\mathcal{A}=\{(\tau_i,a_i,m_i)\}_{i=1}^{N}\), where \(a_i\) is an end-effector command and \(m_i\) specifies valid components. Physical time \(\tau=0\) denotes the current observation. Let \(c\) denote known execution information, including the current end-effector anchor and validity information. An encoder \(E_\phi\) maps the sequence to a latent function \(Z\), and a decoder \(D_\psi\) queries it at physical time \(\tau\):

\begin{equation}
Z=E_\phi(\mathcal{A},c),\qquad a(\tau)=D_\psi(Z,\tau,c).
\end{equation}

The latent has a fixed size, while the timestamps used to observe or query the motion can vary.

The implementation uses an external time coordinate \(u\) offset from the physical clock by \(\Delta t\), the delay between the current observation and the first action target. The physical-time notation above corresponds to

\begin{equation}
\begin{aligned}
\tau&=u+\Delta t,\\
D_\psi(Z,\tau,c)&=D_\psi^{\mathrm{impl}}(Z,\tau-\Delta t,c).
\end{aligned}
\end{equation}

The encoder \(E_\phi\) above also denotes a physical-time interface: its implementation receives external labels \(u_i=\tau_i-\Delta t\). Encoder timestamps and decoder queries are each converted to physical time once at their respective module boundaries. The current-state anchor therefore corresponds to external query \(u=-\Delta t\). The execution segment spans 1.2 s of physical time.

\textbf{Mixture-of-Transformers for Action Function Encoding.} The MoT encoder brings position (XYZ), orientation (Q), and gripper (G) components into a shared latent function representation. In the dual-arm implementation, these streams contain six position coordinates, eight quaternion coordinates, and two gripper commands. Each stream combines the corresponding components of both arms. Input tokens incorporate timestamps and validity masks, so the encoder can process different temporal views without treating missing observations as valid commands.

The encoder first aggregates trajectory tokens into learned latent queries through typed Perceiver blocks. Latent MoT blocks then exchange information across the three action-component streams. Each stream has its own query, key, value, and output projections, normalization layers, and feed-forward networks. Attention is computed jointly over all streams, allowing the specialized Transformer parameters to model each action component while shared attention captures their temporal coordination. The bottleneck contains 96 tokens of width 256, with 32 tokens allocated to each modality.

\textbf{Independent time queries.} The decoder forms a query from the requested physical time, its Fourier features, a modality embedding, and known execution information. Typed cross-attention blocks read the latent memory and produce XYZ, Q, and G outputs. Each requested timestamp is evaluated through its own query-to-memory attention. Consequently, an action at a given timestamp does not depend on the number or ordering of the other requested timestamps. A single latent can therefore produce actions at different execution frequencies through the corresponding time queries.

\textbf{Anchored geometric readout.} Physical time zero corresponds to the observed current state. For each valid arm, the position readout uses the current position \(p_0\) as its anchor:

\begin{equation}
p(\tau)=p_0+\widetilde p(\tau)-\widetilde p(0),
\end{equation}

where \(\widetilde p\) is the decoder's position output before anchoring, expressed on the physical clock. Let \(r(\tau)\) be its normalized quaternion output and \(q_0\) the current orientation. The orientation readout is

\begin{equation}
q(\tau)=\operatorname{normalize}\!\left(q_0\otimes r(0)^{-1}\otimes r(\tau)\right),
\end{equation}

where \(\otimes\) denotes quaternion multiplication. These readouts express predicted motion relative to the observed pose and recover that pose at time zero. The hard anchor applies to position and orientation. Gripper commands use a separate time-indexed knot readout: the decoder predicts gripper values at native knots, and a query selects the nearest knot. This accommodates discrete gripper changes alongside time-queryable pose commands.

\subsection{Learning Cross-Embodiment Dynamics Priors}\label{learning-cross-embodiment-dynamics-priors}\label{sec:pretraining}

Stage A learns reusable dynamics priors from action trajectories. Its inputs are timestamped actions and known execution information, allowing action data to contribute to pretraining without paired language instructions or camera observations. The MoT encoder integrates XYZ/Q/G streams through specialized Transformer parameters and shared attention. This common action organization, together with validity masks and execution context, provides the representational interface for incorporating trajectories from different embodiments.

\textbf{Learning across observation frequencies.} For each trajectory, we construct two observations \(\mathcal{A}^{(1)}\) and \(\mathcal{A}^{(2)}\) at different sampling rates. Both describe the same underlying motion on the same physical clock. The shared encoder produces

\begin{equation}
Z^{(1)}=E_\phi(\mathcal{A}^{(1)},c),\qquad Z^{(2)}=E_\phi(\mathcal{A}^{(2)},c).
\end{equation}

Each latent receives reconstruction supervision at its source timestamps and on an additional query grid. The grid contains 73 points at 60 Hz over external coordinates \(u\in[0,1.2]\) s, mapped to physical time by \(\tau=u+\Delta t\).

Let \(\Delta p\) denote the six-dimensional position error of both arms, \(\Delta g_L\) and \(\Delta g_R\) their gripper errors, and \(\theta_L\) and \(\theta_R\) their shortest rotation errors on \(\mathrm{SO}(3)\). For each arm, \(\theta=\|\log(q^{-1}\hat q)\|_2\) is the geodesic angle in radians. For view \(v\) and query subset \(S\), reconstruction is

\begin{equation}
\begin{aligned}
E_S^{(v)}=\operatorname{mean}_{m_S^{(v)}}\Bigg[&\frac{\|\Delta p\|_2^2}{6}+\frac{\theta_L^2+\theta_R^2}{2}\\
&+\frac{\Delta g_L^2+\Delta g_R^2}{2}\Bigg].
\end{aligned}
\end{equation}

The mean uses the subset's own Boolean mask over batch and query indices. Position is measured in meters, and the three terms use their native scales. Source and grid losses are reduced separately and assigned equal weight. The two-view objective is

\begin{equation}
\begin{aligned}
\mathcal{L}_A={}&\frac{1}{2}\sum_{v=1}^{2}\left(E_{\mathrm{source}}^{(v)}+E_{\mathrm{grid}}^{(v)}\right)\\
&+0.1\operatorname{MSE}(Z^{(1)},Z^{(2)}).
\end{aligned}
\end{equation}

The latent MSE averages over batch, tokens, and latent dimensions. Reconstruction teaches the decoder how motion evolves over time; consistency aligns the representations obtained from different observation frequencies. Together, they establish a shared function space that Stage B can address directly.

\textbf{A scalable trajectory interface.} Separating action-function pretraining from task-specific observation-to-action learning makes trajectory data usable through a common representation. Additional sources enter through timestamped end-effector trajectories, component validity, and known execution information. Shared attention in the MoT encoder captures coordination among action components, while the explicit context preserves information needed to interpret the motion for a particular robot. This is the mechanism through which CoAWM connects cross-embodiment data collection to the scaling of dynamics-prior learning.

\subsection{Connecting the Function Space to a World Model}\label{connecting-the-function-space-to-a-world-model}

Stage B connects the pretrained function space to LaWAM's visual world prediction. Current RGB observations from the head and two wrist cameras, together with a language instruction, are processed by the VLM to produce features \(h_{\mathrm{vlm}}\) and the original latent-action prediction \(\hat a_{\mathrm{latent}}\). The current images also yield LAM visual tokens \(h_t\). The retained world decoder predicts future visual tokens:

\begin{equation}
\hat h_{t+1}=\operatorname{LAMDecoder}(h_t,\hat a_{\mathrm{latent}}).
\end{equation}

Here \(\hat a_{\mathrm{latent}}\) drives the original world prediction, while \(\hat Z\) below is the action-function latent decoded into robot commands. The head combines projected VLM features, predicted future visual tokens, and proprioceptive state \(s\) as its cross-attention memory:

\begin{equation}
M=\operatorname{concat}(P_vh_{\mathrm{vlm}},P_f\hat h_{t+1},P_ss).
\end{equation}

Learned latent queries attend to \(M\) through the head's Transformer decoder. Let \(H\) denote the resulting hidden tokens after LayerNorm. Current LAM visual tokens enter a separate attention residual, with \(H\) as queries and projected current tokens as keys and values:

\begin{equation}
\hat Z=W_{\mathrm{out}}H+W_{\mathrm{res}}\operatorname{Attn}(H,P_ch_t,P_ch_t).
\end{equation}

The residual output projection \(W_{\mathrm{res}}\) is initialized to zero. The head thus combines task semantics and predicted visual evolution in its main memory, then incorporates current visual information at the function-latent output. Proprioception enters the head memory; the decoder's execution information \(c\) remains a separate interface for time-query decoding and anchoring.

\textbf{Expert supervision in function space.} During training, the frozen action encoder maps an expert future trajectory \(\mathcal{A}^{*}\) to a target latent:

\begin{equation}
Z^{*}=\operatorname{sg}\!\left[E_\phi(\mathcal{A}^{*},c)\right],
\end{equation}

where \(\operatorname{sg}\) stops gradients. For each batch sample \(b\), we flatten the full predicted and target latents as \(x_b=\operatorname{vec}(\hat Z_b)\) and \(y_b=\operatorname{vec}(Z_b^{*})\). The latent objective is

\begin{equation}
\mathcal{L}_{\mathrm{latent}}=\operatorname{mean}_b\left[1-\frac{x_b^\top y_b}{\|x_b\|_2\|y_b\|_2}\right].
\end{equation}

The alignment objective computes cosine similarity from the predicted and target latents, and the action decoder directly receives the predicted latent \(\hat Z\).

\textbf{Supervision through decoded actions.}

Using position, rotation-angle, and gripper errors as defined in Stage A, decoded-action supervision uses the scales \(s_p=0.01\) m, \(s_q=0.05\) rad, and \(s_g=0.05\):

\begin{equation}
\begin{aligned}
\mathcal{L}_{\mathrm{phys}}=\frac{1}{3}\Big[&\operatorname{mean}\left((\Delta p/s_p)^2\right)\\
&+\operatorname{mean}\left((\theta/s_q)^2\right)\\
&+\operatorname{mean}\left((\Delta g/s_g)^2\right)\Big].
\end{aligned}
\end{equation}

Stage B optimizes

\begin{equation}
\mathcal{L}_B=\mathcal{L}_{\mathrm{latent}}+0.1\mathcal{L}_{\mathrm{phys}}.
\end{equation}

Latent alignment connects the prediction to the pretrained function representation, while action supervision ties it to the commands produced at native target times. The function encoder and decoder remain frozen. Gradients through the decoder train the latent prediction pathway and backbone adapters, preserving the function space while adapting its use to the task. The retained world-prediction branch supplies features to the head and remains frozen.

\subsection{Time-Query Action Execution}\label{time-query-action-execution}

At each replanning step, CoAWM uses the current observation to predict one latent function and anchors its pose readout to the observed current state.

For frequency \(f\), we query the nominal endpoint of each command interval. For \(k=0,\ldots,fT_{\mathrm{exec}}-1\), where \(T_{\mathrm{exec}}\) denotes the execution-segment duration, the physical query and corresponding external decoder input are

\begin{equation}
\begin{aligned}
\tau_k&=\frac{k+1}{f},\qquad u_k=\frac{k+1}{f}-\Delta t,\\
a_k&=D_\psi(\hat Z,\tau_k,c).
\end{aligned}
\end{equation}

Native execution uses 30 Hz.

The inference path consists of observation encoding, visual world prediction, latent-function prediction, and time-query decoding. The expert trajectory and function encoder are used only for training. This organization connects pretrained dynamics priors to task-aware action generation and frequency-specific execution through a shared function interface.

\section{Experiments}
We evaluate continuous action functions from four perspectives: representation accuracy, closed-loop execution, cross-frequency querying, and action quality. We first examine whether the function space can accurately represent robot actions, then evaluate task performance after alignment with LaWAM, focusing on the effect of the continuous representation on the smoothness of initial execution. (These experiments provide a rapid validation of the method; a subsequent arXiv version will expand the training scale and perform full fine-tuning.)

\subsection{Representation Accuracy of Continuous Actions}
\begin{figure}[!t]
\centering
\includegraphics[width=\columnwidth]{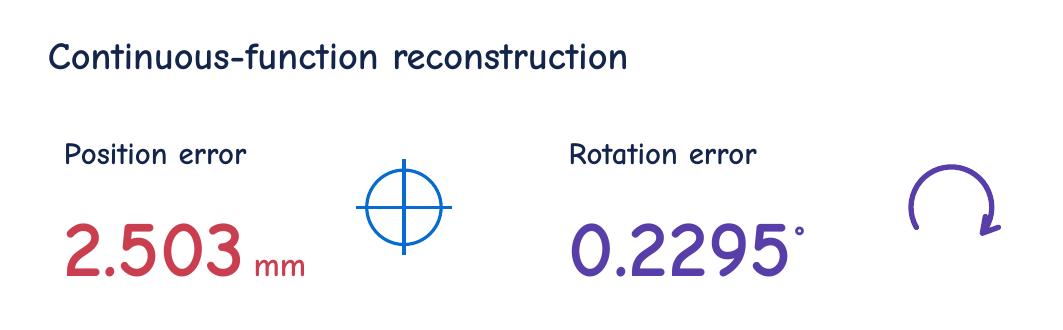}
\caption{Mean position and rotation reconstruction errors of the function autoencoder on 8,192 held-out evaluation windows used for model selection.}
\label{fig:e01}
\end{figure}

A continuous function representation preserves action details while providing a unified query interface for different time points. On 8,192 held-out evaluation windows, the function autoencoder achieves a position reconstruction error of 2.503~mm and a rotation error of $0.2295^\circ$. This result shows that the function space can represent the original action trajectories with high accuracy, providing a representation foundation for subsequent world-model alignment (Fig.~\ref{fig:e01}).

\subsection{Smoother Initial Execution at Similar Movement Speeds}
\begin{figure*}[!t]
\centering
\includegraphics[width=\textwidth]{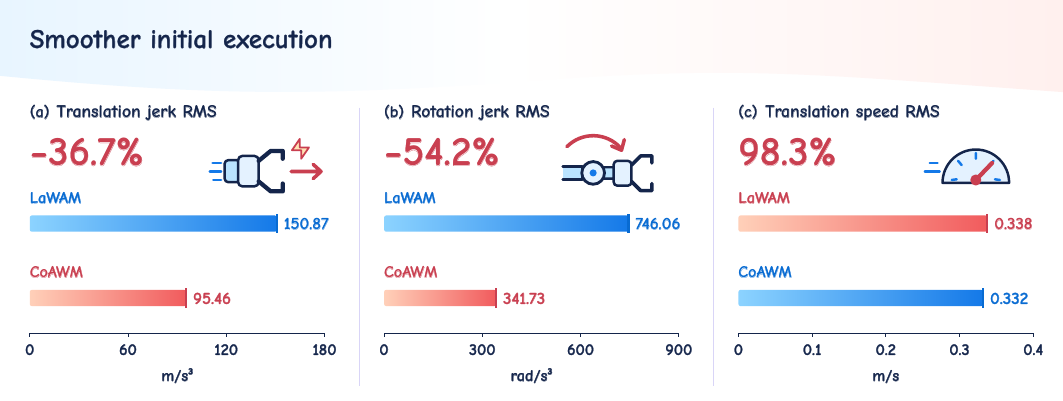}
\caption{Initial execution quality during the first 1~s across 800 paired trials in 16 tasks. The highlighted percentages show the reduction in translation jerk RMS, the reduction in rotation jerk RMS, and the retained translation speed RMS relative to LaWAM, respectively. Metrics describe actual end-effector motion, with equal task weighting and both arms pooled.}
\label{fig:e11-main}
\end{figure*}

Continuous action functions significantly reduce abrupt changes in motion during initial execution. Across 16 tasks and 800 pairs of matched initial conditions, we measure motion metrics over the first second of each execution. Compared with LaWAM, position jerk RMS decreases from 150.87 to 95.46, a reduction of 36.7\%; rotation jerk RMS decreases from 746.06 to 341.73, a reduction of 54.2\%. The velocity RMS values of the two methods are 0.338 and 0.332, respectively, showing that continuous action functions reduce abrupt changes in motion without reducing action execution speed (Fig.~\ref{fig:e11-main}).

This result demonstrates the improvement in initial execution quality provided by continuous action representations. In the same evaluation, our method achieves a task success rate of 75.1\%. However, we find that substantial jerk tails remain over longer execution windows, and plan to further address action stability during sustained execution and replanning in subsequent optimization.

\subsection{Multitask Manipulation and Cross-Frequency Execution}
The aligned function policy can perform a variety of bimanual manipulation tasks. In evaluations with 50 trials per task, it achieves a success rate of 94\% on the two-bottle task, 90\% on both left and right A2B tasks, 88\% on the burger task, and 86\% on the hammer task. In a separate preliminary evaluation covering 50 tasks with three trials per task, both methods complete all three trials on 30 tasks.

\begin{figure*}[!t]
\centering
\includegraphics[width=\textwidth]{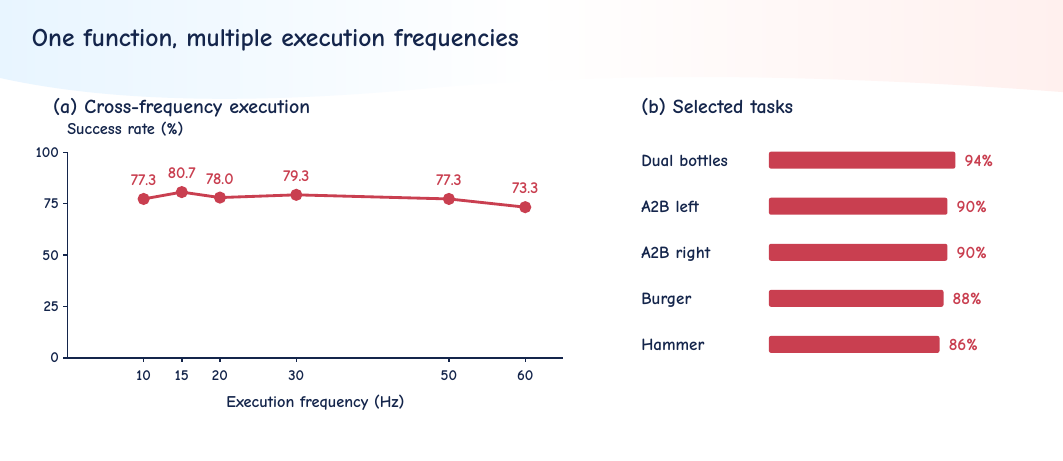}
\caption{(a) CoAWM success rates at six execution frequencies, evaluated with 150 trials per frequency across 50 tasks (three trials per task), with the execution segment and replanning period fixed at 1.2~s. (b) Five selected tasks from a separate 16-task evaluation with 50 trials per task. The panels therefore use different evaluation cohorts and denominators.}
\label{fig:e10}
\end{figure*}

By changing the time query points, the same function policy can output actions at 10, 15, 20, 30, 50, and 60~Hz while keeping the execution segment and replanning period fixed at 1.2~s. The corresponding numbers of successes are 116, 121, 117, 119, 116, and 110, with 150 trials evaluated at each frequency. These results demonstrate cross-frequency execution within a fixed physical-time window (Fig.~\ref{fig:e10}).

\subsection{Reducing Execution Energy Consumption}
\begin{figure*}[!t]
\centering
\includegraphics[width=\textwidth]{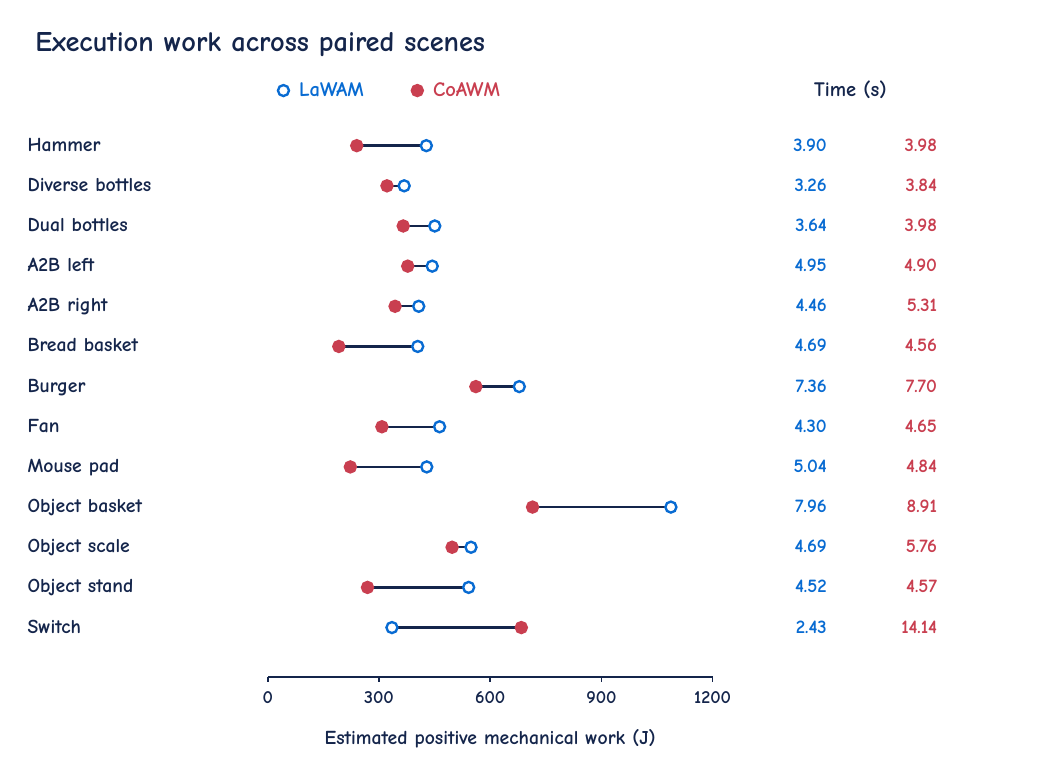}
\caption{Estimated positive mechanical work for the 13 jointly successful scenes among 16 prespecified scenes, computed from PD torque estimates and measured joint velocities. Open blue points denote LaWAM and filled red points denote CoAWM; connecting segments identify paired scenes. The two time columns report physical execution duration in seconds using the same method colors.}
\label{fig:e12}
\end{figure*}

We further use positive mechanical work based on PD torque estimates to measure energy consumption during execution. Among 16 prespecified scenarios, on the 13 scenarios in which both methods succeed, the function policy reduces the mean estimated positive mechanical work from 506.66 to 392.13, a reduction of 22.6\%, with lower values in 12 scenarios (Fig.~\ref{fig:e12}).

\section{Discussion}
Continuous function representations separate the learning of action structure from the choice of execution frequency. The model learns how actions evolve over physical time, while the execution side obtains commands at the required frequency through time queries. Execution frequency can therefore be adjusted while keeping the action-segment duration and replanning period unchanged, allowing the same action representation to serve different execution clocks.

This representation also makes action quality an important evaluation dimension alongside task success rate. The current experiments demonstrate high trajectory reconstruction accuracy and smoother initial execution at similar movement speeds. At the same time, offline reconstruction quality, closed-loop task success rate, and stability during sustained execution have different optimization requirements. Considering these metrics together enables a more complete evaluation of the practical effects of aligning function representations with a world model.

The shared function space provides a unified interface for expanding the training scale. During function pretraining, we will introduce action trajectories from more embodiments and collection frequencies; during world-model alignment, we will further perform full fine-tuning, using visual predictions and task information to learn task-related action functions. Subsequent experiments will examine how larger-scale data and training affect task performance, cross-embodiment transfer, and stability during sustained execution.

\section{Conclusion}
We present CoAWM, which connects action representation learning, world-model prediction, and time-query execution through a shared function space. Two-stage training uses the spatiotemporal structure in trajectories for task-related action generation and supports different execution frequencies through time queries. Preliminary experiments demonstrate the reconstruction accuracy, bimanual task execution capabilities, and improvements in initial action quality of this representation, providing a foundation for further expanding the scale of data and training.

\clearpage
{\small
\bibliographystyle{ieeenat_fullname}
\bibliography{main}
}
\end{document}